# Bilaterally Mirrored Movements Improve the Accuracy and Precision of Training Data for Supervised Learning of Neural or Myoelectric Prosthetic Control

Jacob A. George, *Student Member, IEEE,* Troy N. Tully, Paul C. Colgan, and Gregory A. Clark

*Abstract*—Intuitive control of prostheses relies on training algorithms to correlate biological recordings to motor intent. The quality of the training dataset is critical to run-time performance, but it is difficult to label hand kinematics accurately after the hand has been amputated. We quantified the accuracy and precision of labeling hand kinematics for two different approaches: 1) assuming a participant is perfectly mimicking predetermined motions of a prosthesis (mimicked training), and 2) assuming a participant is perfectly mirroring their contralateral hand during identical bilateral movements (mirrored training). We compared these approaches in non-amputee individuals, using an infrared camera to track eight different joint angles of the hands in real-time. Aggregate data revealed that mimicked training does not account for biomechanical coupling or temporal changes in hand posture. Mirrored training was significantly more accurate and precise at labeling hand kinematics. However, when training a modified Kalman filter to estimate motor intent, the mimicked and mirrored training approaches were not significantly different. The results suggest that the mirrored training approach creates a more faithful but more complex dataset. Advanced algorithms, more capable of learning the complex mirrored training dataset, may yield better run-time prosthetic control.

## I. INTRODUCTION

Even though the physical hand is missing after an amputation, most transradial amputees still retain the neural circuits and much of the musculature that control the hand. Electromyographic (EMG) signals, recorded from the extrinsic hand muscles still present in the residual limb, can be used to intuitively control prostheses.

Algorithms are typically trained to decode motor intent from EMG under a supervised-learning paradigm that involves a dataset consisting of EMG and labeled kinematics. The quality of this dataset is critically important in developing robust and accurate control algorithms. Ideally, this dataset would be generated by simultaneously recording EMG from the extrinsic muscles of the hand and recording kinematics from the fingers of the hand. However, after the hand is amputated, there is no direct way to determine motor intent and correctly label hand kinematics.

Traditionally, motor intent is determined by assuming the amputee participant is perfectly mimicking the predetermined motion of a prosthesis with their missing hand (i.e., mimicked training) [1]–[3]. However, the validity of this assumption is unclear. There is at least some uncertainty in the temporal alignment of the predetermined and mimicked motions due to participant reaction time; algorithms often preprocess the training data by aligning the preprogrammed kinematics with EMG features in order to account for temporal delays [2], [3].

We hypothesized that, in addition to temporal delays, the mimicked training approach would also not account for variations in kinematic amplitude (i.e., the degree of flexion or extension), biomechanical coupling, and changes in resting hand position over time. To this end, here we precisely quantify these potential sources of error in intact individuals where the actual hand kinematics can be relatively accurately determined. We also compare the errors associated with the traditional mimicked training approach to an alternative training approach that assumes a unilateral amputee participant is perfectly mirroring their intact contralateral limb during synchronized bilateral movements (i.e., mirrored training) [4]. Furthermore, we directly compare the ability to estimate motor intent, with a modified Kalman filter [3], using training data gathered under the mimicked and mirrored training approaches. These results can be used to generate more accurate training datasets, and therefore constitute an important step towards dexterous bionic arms.

## II. METHODS

### A. Human Subjects

A total of seven non-amputee human participants were used in this study. All participants were between the ages of 18 and 30. Informed consent and experimental protocols were carried out in accordance with the University of Utah Institutional Review Board.

### B. Experimental Setup

Participants were instructed to mimic preprogrammed movements of a virtual hand (Modular Prosthetic Limb, MSMS; Johns Hopkins Applied Physics Lab, Baltimore, MD) displayed on a computer monitor. Participants mimicked the virtual right hand with both their right and left hands simultaneously. An infrared hand-tracking device (Leap Motion; Ultrahaptics, San Francisco, CA) was placed approximately eight inches below their hands to track the motion of their fingers and wrist. EMG from the right forearm was recorded in synchrony with hand kinematics using a custom sleeve with embedded surface electrodes. Three different kinematic signals were recorded simultaneously: 1) True Kinematics – from the intact right hand, 2) Contralateral Kinematics – from the intact left hand, and 3) Virtual Kinematics – from the virtual right hand (Fig. 1).

### C. Signal Acquisition

Infrared hand images were converted to joint angle using custom MATLAB software. A total of eight joint angles were calculated for each hand: D1 abduction/adduction, D1-D5 flexion/extension, wrist flexion/extension, and wrist pronation/supination. EMG was recorded from 32 single-ended surface electrodes embedded in a custom neoprene sleeve. EMG recordings were sampled at 1 kHz and filtered using the Grapevine Neural Interface Processor (Ripple Neuro LLC, Salt Lake City, UT, USA) as described in [3]. EMG features used for estimating motor intent consisted of the 300-ms smoothed mean absolute value on 528 channels (32 single-ended channels and 496 calculated differential pairs), calculated at 30 Hz [3].

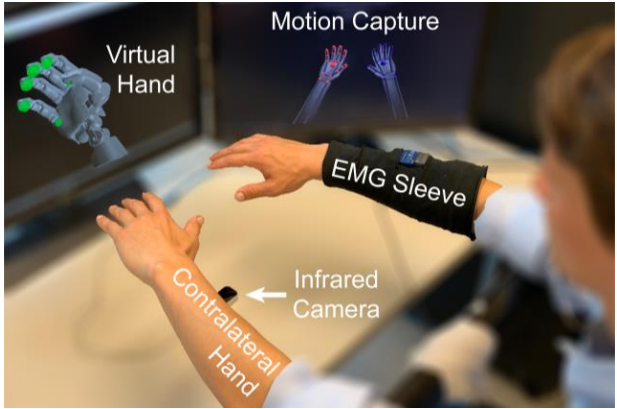

Figure 1: Experimental setup. Healthy participants were instructed to mimic the preprogrammed movements of a virtual right hand with both their right and left hands simultaneously. An infrared-camera-based motion-capture device was used to track the kinematics of the participants' hands. Electromyography (EMG) was recorded from the right forearm using surface electrodes embedded in a neoprene sleeve.

### D. Training Movements

Participants were instructed to mimic the movements of the virtual hand (Virtual Kinematics) with their right hand (True Kinematics) and left hand (Contralateral Kinematics) simultaneously. These hand movements included individuated movements of each DOF of the virtual prosthetic hand (flexions/extensions of D1-D5; wrist flexion/extension; wrist pronation/supination; thumb abduction/adduction) as well as two combination movements (simultaneous flexion of D1-D5; simultaneous extension of D1-D5), for a total of 20 unique movements. The participants performed 10 trials of each movement, for a total of 200 trials. All 10 trials of each movement were performed sequentially, one after another, and the total duration of each individual movement was 1.5 s (made up of a 0.7-s deviation away from the resting hand position, a 0.1-s hold-time at the position of maximum deviation, and then another 0.7-s deviation back to the original resting hand position). There was a 1-s intertrial interval and a 30-s resting period before the start of the first trial (to assess resting hand posture).

### E. Comparison of Virtual (Instructed) and Intact (Actual) Hand Movements

Preprogrammed movements of the virtual hand are perfectly consistent and isolated, ignoring the variability and biomechanical coupling associated with endogenous hand movements. To this end, we quantified the amount of biomechanical coupling and drift in the resting hand position throughout the data collection process using the True Kinematics.

*1) Biomechanical coupling.*

We estimated and operationally defined biomechanical coupling as the unintended movement of non-target DOFs when attempting to move a target DOF. For example, the virtual hand would perform D4 extension perfectly isolated such that no other DOFs move. However, when the participants attempted to perform isolated D4 extension, there was often associated movement on D3 and/or D5. Biomechanical coupling was quantified as the peak deviation from the resting position of non-target DOFs, where the resting position was defined as the mean value during the previous intertrial interval.

*2) Resting-hand-position drift.*

We defined resting-hand-position drift as the changes in resting position of the hand throughout the entire data collection process. Drift was quantified at each intertrial interval as the current resting position at that intertrial interval to the resting position recorded during the 30 s prior to data collection.

### F. Comparison of Mimicked Training and Mirrored Training

We hypothesized that participants would not be able to recreate the precision of the virtual hand when attempting to mimic the preprogrammed movements. To this end, we quantified the differences in the magnitude and timing of the movements between the True Kinematics and the Contralateral Kinematics as well as between the True Kinematics and the Virtual Kinematics. Likewise, we quantified the differences in the magnitude variance and timing variance.

*1) Magnitude of movements*

We defined the magnitude of movements as the maximum deviation away from the resting hand position. For example, the virtual hand would perform 10 trials of D4 extension such that each trial had the exact same maximum deviation. However, when the participants attempted to perform D4 extension, there was often variations in the maximum deviation. For each trial, we calculated the error in magnitude as the difference in maximum deviation of the True Kinematics relative to the Virtual Kinematics (mimicked training) or relative to the Contralateral Kinematics (mirrored training).

*2) Timing of movements*

For each trial, the difference in timing was quantified as the difference in the time at which the maximum deviation occurred for the True Kinematics relative to when it occurred for the Virtual Kinematics (mimicked training) or to when it occurred for the Contralateral Kinematics (mirrored training).

### G. Comparison of Run-Time Estimates of Motor Intent

The EMG activity recorded during this task, as well as the Virtual or Contralateral Kinematics, served as training data for a modified Kalman filter (MKF) to estimate motor intent [3]. Two MKFs were trained: 1) using the Virtual Kinematics (mimicked training) and 2) using the Contralateral Kinematics (mirrored training).

To avoid complications due to the participant's reaction time for mimicked training only, we aligned the kinematics with the EMG by shifting the kinematic positions by a lag that was determined by maximizing the cross-correlation. This alignment was performed uniformly across all trials [3].

The algorithms were trained using the same random 50% of the trials for each movement. The remaining 50% of the trials were used to evaluate the performance (root-mean-squared error; RMSE) of the algorithms under two conditions: 1) the ability to recreate the training data (i.e., Virtual Kinematics or Contralateral Kinematics), and 2) the ability to recreate the True Kinematics. Improvements in the second metric would ultimately yield more dexterous and intuitive prosthetic control. Alignment between the first and second metrics would indicate that improvements in algorithm performance offline are likely to translate to improvements online.

*H. Statistical Analyses*

The median values for each participant were analyzed such that the total number of samples was equal to the number of participants ($N = 7$). Outliers in the performance metrics (more than 1.5 interquartile ranges above the upper quartile or below the lower quartile) were removed from the data prior to statistical analyses. One-sample t-tests were performed to determine if the biomechanical coupling and resting-hand-position drift associated with True Kinematics were statistically non-zero (e.g., different from the Virtual Kinematics). Two-sample paired t-tests were used to compare between mimicked training and mirrored training for all other metrics.

III. RESULTS

*A. Preprogrammed movements of a virtual hand did not account for biomechanical coupling or temporal changes in resting hand position*

Using an infrared motion-capture device, we quantified the deviations in the True Kinematics due to biomechanical coupling and temporal changes in resting hand position. We found significantly non-zero kinematic deviations for both ($p$'s < 0.001). Biomechanical coupling resulted in 11.43 ± 0.57% (mean ± S.E.M.) deviation in the recorded kinematics and resting-hand-position drift resulted in 7.07 ± 0.56% deviation (Fig. 2).

*B. Mirroring contralateral movements reduced the error and variability of movement magnitude*

We compared the ability of participants to accurately mimic the preprogrammed movements of a virtual hand or mirror their own contralateral movements during identical bilateral movements. We found that the magnitude of movements for the True Kinematics was significantly more closely related to that of the Contralateral Kinematics than to that of the Virtual Kinematics (6.67 ± 0.42% vs 12.89 ± 1.35%, $p < 0.005$). Furthermore, the variance in magnitude errors was significantly less for the Contralateral Kinematics than for the Virtual Kinematics (0.53 ± 0.07% vs 1.45 ± 0.18%, $p < 0.005$; Fig. 3).

*C. Mirroring contralateral movements reduced the error in movement timing, but increased the variability of errors*

We found that errors in the timing of movements were also significantly less for the Contralateral Kinematics than for the Virtual Kinematics (0.03 ± 0.02 s vs 0.08 ± 0.02 s, $p < 0.05$). However, the variance of timing errors was significantly greater for the Contralateral Kinematics than for the Virtual Kinematics (0.06 ± 0.01 s vs 0.05 ± 0.01 s, $p < 0.005$; Fig. 3).

*D. True Kinematics were more closely aligned with Contralateral Kinematics than with Virtual Kinematics*

Overall, the RMSE between True Kinematics and Contralateral Kinematics was significantly lower than the RMSE between True Kinematics and Virtual Kinematics (0.16 ± 0.01% vs 0.19 ± 0.01%, $p < 0.05$; Fig. 3).

*E. Datasets gathered under the mirrored training approach were more complex and difficult for algorithms to recreate, but offline analyses of these datasets may be more faithful to a participant's intent*

We trained two different MKFs to estimate motor intent from EMG activity recorded in synchrony with hand kinematics. We found that the MKF trained under the mirrored training approach (i.e., using Contralateral Kinematics) was significantly worse at recreating the training data than the MKF trained under the mimicked training approach (i.e., using Virtual Kinematics); the RMSE between the kinematic predictions and the training data kinematics was significantly greater for the mirrored training MKF (0.15 ± 0.01% vs 0.09 ± 0.01%, $p < 0.001$). However, there was no significant difference in the RMSE between the kinematic predictions and the True Kinematics ($p = 0.44$; Fig. 4).

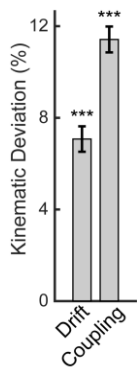

Figure 2: Deviation in kinematics due to biomechanical coupling and resting-hand-position drift. Kinematic deviations were significantly non-zero for both coupling and drift. Data show mean ± S.E.M. *** $p < 0.001$, one-sample t-test.

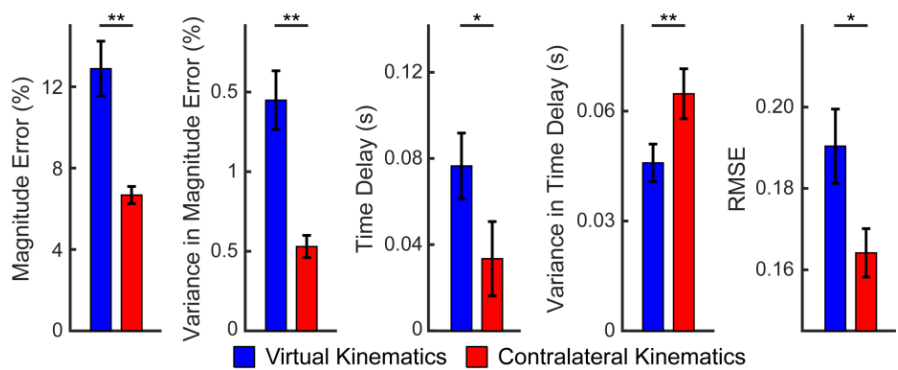

Figure 3: Errors associated with assuming an individual is perfectly mimicking preprogrammed movements of a virtual hand (Virtual Kinematics) or assuming an individual is perfectly mirroring their contralateral limb (Contralateral Kinematics). The True Kinematics were more closely aligned with the Contralateral Kinematics than were the Virtual Kinematics. Contralateral Kinematics had lower Root Mean Squared Error (RMSE) and lower errors associated with movement magnitude and movement timing. Contralateral kinematics also had lower variance in magnitude errors but had higher variance in timing errors. Data show mean ± S.E.M. * $p < 0.05$, ** $p < 0.01$, two-sample paired t-test.

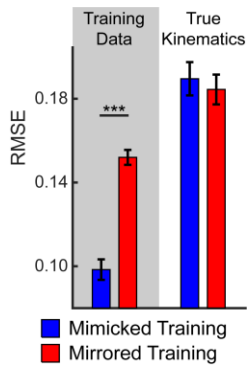

Figure 4: Accuracy of kinematic predictions. Two modified Kalman filters (MKF) were trained, one under the mimicked training approach and one under the mirrored training approach. The mimicked training MKF was significantly better at recreating the training than was the mirrored training MKF. There was no significant difference between the MKFs when attempting to recreate the True Kinematics. Data show mean ± S.E.M. *** $p <$ 0.001, two-sample paired t-test.

## IV. Discussion

Accurately labeled training data is critically important for algorithm performance. Here, we compared the accuracy of two different approaches to label motor intent for prosthetic control algorithms. Overall, we found that labeling hand kinematics with bilaterally mirrored movements is more accurate and precise than is mimicking preprogrammed virtual movements, although these benefits have not yet been translated into improved prosthetic control for recreating the True Kinematics and presumed user intent.

For four of the five metrics we analyzed, the Contralateral Kinematics were better than the Virtual Kinematics. Overall, the RMSE was less between the True Kinematics and the Contralateral Kinematics than between the True Kinematics and the Virtual Kinematics, likely due to biomechanical coupling that was observed for both the right and left hands but not for the virtual hand. In addition, there was a 48% reduction in errors associated with movement magnitude for the Contralateral Kinematics. To this end, the participants demonstrated superior accuracy and precision when mirroring the magnitude of their own hand movements than when mimicking the magnitude of virtual hand movements.

With respect to errors in timing, the Contralateral Kinematics were more accurate, but less precise than the Virtual Kinematics. This is likely attributed to the fact there is a visual reaction time associated with mimicking preprogrammed movements of a virtual hand, and this leads to a large but consistent delay in timing. In contrast, bilaterally mirrored movements are more temporally aligned, but minor inconsistencies can be seen in both directions (i.e., the Contralateral Kinematics can either precede or lag behind the True Kinematics).

Taken together, the overall results generally suggest that the mirrored training approach is a better way to label motor intent. Importantly from a practical perspective, however, the MKF using the mimicked training approach was better at recreating the training data than was the MKF using the mirrored training, and was as good at recreating the true kinematics. Why didn't the potential benefits of mirrored training translate to improved MKF decodes? We propose that the mirrored training approach results in a more faithful, but much more complex dataset. In contrast, the Virtual Kinematics provide a relatively simple dataset that can be easily learned and recreated. When the training dataset is complicated by biomechanical coupling and temporal changes in resting hand position, the performance of the MKF degrades. The MKF performance was similar for recreating the mirrored training dataset and the True Kinematics, suggesting that offline performance on the mirrored training dataset may be more indicative of online performance.

Ultimately, more complex datasets require more complex algorithms. We hypothesize that deep neural networks capable of capturing non-linear changes in kinematics due to biomechanical coupling or resting-hand-position drift will be able to take advantage of this more complex, but more accurate, mirrored training dataset to improve estimates of motor intent.

There is some error associated with the infrared hand-tracking device, although it was not measured here. We propose that algorithms trained on Contralateral Kinematics should be trained to recreate a confidence interval of kinematics instead of an absolute value. For this reason, dataset aggregation [5] may be better suited for mirrored training.

Future work should also validate the performance of algorithms online, through functional activities of daily living. There may be benefits to exploiting the full capabilities of bionic arms, and these could be realized by excluding some aspects of endogenous hand kinematics from the training data. Contrastingly, it has been shown that more faithful and biomimetic motor control can enhance prosthesis embodiment [6]–[8]. When coupled with biomimetic sensory feedback [9], [10], prosthetics may be able to recreate the physical and psychological experience of the human hand.

## V. Conclusion

In order to train algorithms to accurately recreate motor intent, we need to accurately identify and label of motor intent. Here, we demonstrate that when the physical limb is missing, motor intent is best determined by tracking the motion of the contralateral limb while the participant performs bilaterally mirrored movements. This approach captures the complex hand kinematics that arise from biomechanical coupling and temporal drifts in hand posture. Algorithms that are able to learn this complex dataset will likely yield more dexterous and biomimetic prosthetic control.

## Author Contributions

J.A.G. designed experiments, oversaw data collection, performed data analysis and wrote the manuscript. T.N.T. developed software, performed experiments and assisted with data analysis. P.C.C. assisted with software development. G.A.C. oversaw all aspects of the research. All authors contributed to the revision of the manuscript.

## Acknowledgment

This work was funded by: DARPA, BTO, Hand Proprioception and Touch Interfaces program, Space and

REFERENCES

[1] J. Nieveen et al., "Polynomial Kalman filter for myoelectric prosthetics using efficient kernel ridge regression," in *2017 8th International IEEE/EMBS Conference on Neural Engineering (NER)*, 2017, pp. 432–435, doi: 10.1109/NER.2017.8008382.

[2] J. A. George, M. R. Brinton, C. C. Duncan, D. T. Hutchinson, and G. A. Clark, "Improved Training Paradigms and Motor-decode Algorithms: Results from Intact Individuals and a Recent Transradial Amputee with Prior Complex Regional Pain Syndrome," in *2018 40th Annual International Conference of the IEEE Engineering in Medicine and Biology Society (EMBC)*, 2018, pp. 3782–3787, doi: 10.1109/EMBC.2018.8513342.

[3] J. A. George, T. S. Davis, M. R. Brinton, and G. A. Clark, "Intuitive neuromyoelectric control of a dexterous bionic arm using a modified Kalman filter," *J. Neurosci. Methods*, vol. 330, p. 108462, Nov. 2019, doi: 10.1016/j.jneumeth.2019.108462.

[4] T. S. Davis et al., "Restoring motor control and sensory feedback in people with upper extremity amputations using arrays of 96 microelectrodes implanted in the median and ulnar nerves," *J. Neural Eng.*, vol. 13, no. 3, p. 036001, Jun. 2016, doi: 10.1088/1741-2560/13/3/036001.

[5] H. Dantas, D. J. Warren, S. Wendelken, T. Davis, G. A. Clark, and V. J. Mathews, "Deep Learning Movement Intent Decoders Trained with Dataset Aggregation for Prosthetic Limb Control," *IEEE Trans. Biomed. Eng.*, pp. 1–1, 2019, doi: 10.1109/TBME.2019.2901882.

[6] D. M. Page et al., "Motor Control and Sensory Feedback Enhance Prosthesis Embodiment and Reduce Phantom Pain After Long-Term Hand Amputation," *Front. Hum. Neurosci.*, vol. 12, 2018, doi: 10.3389/fnhum.2018.00352.

[7] A. Kalckert and H. H. Ehrsson, "Moving a Rubber Hand that Feels Like Your Own: A Dissociation of Ownership and Agency," *Front. Hum. Neurosci.*, vol. 6, p. 40, 2012, doi: 10.3389/fnhum.2012.00040.

[8] M. Tsakiris, G. Prabhu, and P. Haggard, "Having a body versus moving your body: How agency structures body-ownership," *Conscious. Cogn.*, vol. 15, no. 2, pp. 423–432, Jun. 2006, doi: 10.1016/j.concog.2005.09.004.

[9] J. A. George et al., "Biomimetic sensory feedback through peripheral nerve stimulation improves dexterous use of a bionic hand," *Sci. Robot.*, vol. 4, no. 32, p. eaax2352, Jul. 2019, doi: 10.1126/scirobotics.aax2352.

[10] G. Valle et al., "Biomimetic Intraneural Sensory Feedback Enhances Sensation Naturalness, Tactile Sensitivity, and Manual Dexterity in a Bidirectional Prosthesis," *Neuron*, vol. 100, no. 1, pp. 37-45.e7, Oct. 2018, doi: 10.1016/j.neuron.2018.08.033.